\pdfoutput=1
\documentclass[11pt]{article}
\usepackage{titlesec}

\usepackage{enumitem} % Add this to your preamble
 
\usepackage[preprint]{acl}

\usepackage{times}
\usepackage{latexsym}
\usepackage{subcaption} % Enables subfigures

\usepackage[T1]{fontenc}
\usepackage[utf8]{inputenc}

\usepackage{microtype}

\usepackage{inconsolata}

\usepackage{graphicx}

\usepackage{pgfplots}
\usepgfplotslibrary{fillbetween}
\usetikzlibrary{patterns}
\usetikzlibrary{pgfplots.groupplots}
\pgfplotsset{compat=1.17}
\usepackage{lscape}
\usepackage{multirow}
\usepackage{dblfloatfix}
\usepackage{footnote}
\usepackage{graphicx}
\usepackage{tikz}
\usepackage{booktabs, tabularx}
\usepackage{amssymb}
\usepackage{amsmath, nccmath}

\usepackage{xurl}

\title{GenKnowSub: Improving Modularity and Reusability of LLMs through General Knowledge Subtraction}

\author{
\textbf{ Mohammadtaha Bagherifard\textsuperscript{2} } \quad
\textbf{Sahar Rajabi\textsuperscript{1}\thanks{\ \ Equal contribution.}} \quad
\textbf{Ali Edalat\textsuperscript{1}\footnotemark[1]} \quad \\
\textbf{Yadollah Yaghoobzadeh\textsuperscript{1,2}} \\
\textsuperscript{1}School of Electrical and Computer Engineering,\\
College of Engineering, University of Tehran, Tehran, Iran \\
\textsuperscript{2}Tehran Institute for Advanced Studies, Khatam University, Tehran, Iran\\
\texttt{taha.bagheri98@gmail.com}, \\  \textnormal{\texttt{\{sahar.rajabi, ali.edalat, y.yaghoobzadeh\}@ut.ac.ir}}
}

\begin{document}
\maketitle
\begin{abstract}
% LLMs often struggle with zero-shot generalization due to the entanglement of general knowledge and task-specific adaptations. In this work, we propose a simple modular framework that disentangles these components by building a library of task-specific LoRA adapters alongside a general knowledge LoRA trained on general-domain data. We subtract the general knowledge from each task-specific adapter to obtain residual modules focusing exclusively on task-relevant information. By reusing these refined adapters and leveraging the Arrow routing algorithm \citep{ostapenko2024towards}, our approach dynamically selects and combines modules for new inputs without additional retraining. 
% Our studies on the Phi-3 model reveal that employing general knowledge LoRAs derived from diverse language sources—across English, French, and German—yields consistent performance gains in both monolingual and crosslingual settings across a wide set of benchmarks. Compared to both the Phi-3 baseline and the standard Arrow, our framework achieves higher accuracy by reducing redundancy and enhancing the precision of task-specific representations, thereby paving the way toward more adaptable, reusable, and cross-lingually robust modular LLMs.

Large language models often struggle with zero-shot generalization, and several modular approaches have been proposed to address this challenge. Yet, we hypothesize that a key limitation remains: the entanglement of general knowledge and task-specific adaptations. To overcome this, we propose a modular framework that disentangles these components by constructing a library of task-specific LoRA modules alongside a general-domain LoRA. By subtracting this general knowledge component from each task-specific module, we obtain residual modules that focus more exclusively on task-relevant information—a method we call general knowledge subtraction (GenKnowSub). Leveraging the refined task-specific modules and the Arrow routing algorithm \citep{ostapenko2024towards}, we dynamically select and combine modules for new inputs without additional training.
Our studies on the Phi-3 model and standard Arrow as baselines reveal that using general knowledge LoRAs derived from diverse languages, including English, French, and German, yields consistent performance gains in both monolingual and cross-lingual settings across a wide set of benchmarks.
Further experiments on Phi-2 demonstrate how GenKnowSub generalizes to weaker LLMs. The complete code and data are available at \url{https://github.com/saharsamr/Modular-LLM}.

% Cross-lingual evaluations further validate that isolating shared linguistic knowledge improves task-specific performance. By disentangling knowledge domains and employing dynamic module routing, our framework enables efficient generalization to new tasks. This scalable solution advances transfer learning, offering a robust approach for adapting LLMs to unseen challenges with minimal computational overhead.

\end{abstract}

\section{Introduction}
The rapid advancement of large language models (LLMs) has led to their widespread adoption in various NLP tasks, ranging from text generation to machine translation and question-answering \citep{NEURIPS2020_1457c0d6, JMLR:v21:20-074}. Despite their remarkable performance, a key challenge remains: ensuring effective generalization to unseen tasks without the need for extensive retraining \citep{bommasani2022opportunitiesrisksfoundationmodels, wei2022emergent}.

In modular zero-shot transfer approaches \citep{pfeiffer2023modular}, a two-stage process is typically followed: (i) task-specific modules are obtained via parameter-efficient fine-tuning (PEFT) methods, such as LoRA \citep{hu2021loralowrankadaptationlarge}, Adapters \citep{pmlr-v97-houlsby19a}, and \(\text{(IA)}^3\) \citep{liu2022fewshot}, on a multitask dataset (ii) a routing function is used to select and combine task-specific modules to address a new task. While some routing functions require joint training alongside task-specific modules \citep{fedus2022switch, caccia2023multihead, ponti-etal-2023-combining}, recent approaches employ post-hoc routing methods that require no further training \citep{chronopoulou-etal-2023-adaptersoup, ostapenko2024towards}. Hybrid approaches also exist, where the routing function is trained separately on a downstream dataset after freezing task-specific modules \citep{Muqeeth2024LearningTR, huang2024lorahub}.

% In this paper, we propose General Knowledge Subtraction or \textit{GenKnowSub}, a method that removes redundant knowledge from task-specific modules to improve generalization to unseen tasks, further advancing the democratization of LLMs. We use LoRA as the PEFT module and adopt Arrow, introduced in \citet{ostapenko2024towards}, as the routing function. To eliminate redundant knowledge from task-specific LoRA modules, we first construct a general knowledge LoRA using a broad corpus and then subtract it from each task-specific module. The Arrow algorithm, a token-based post-hoc routing method that maps each input token to the most relevant task modules without requiring additional training, then dynamically selects and integrates the most relevant LoRAs to handle unseen tasks. An overview of our proposed approach is shown in Figure \ref{fig:approach_overview}.
% \input{images/overview}

\begin{figure*}[h!]
    \centering
    \includegraphics[scale=0.34]{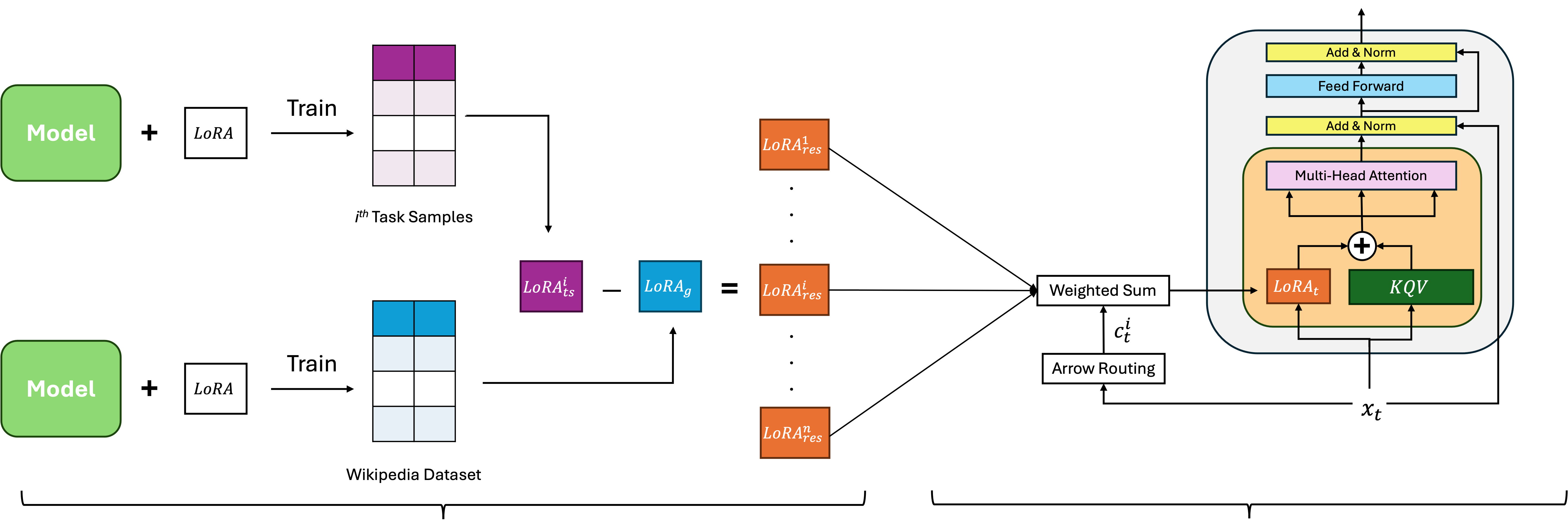}
    
    \vspace{0.3cm} % Adds some space between the image and labels
    \makebox[\linewidth]{ 
    \hfill \small (a) \textit{Training the Modules and General Knowledge Subtraction} \hfill
    \hspace{1.4cm} % Adjust this value for more or less spacing
    \small (b) \textit{Dynamic Task Adaptation via Arrow Routing} \hfill
}
    \caption{Overview of our proposed approach. (a) illustrates the process of training task-specific and general modules, followed by performing general knowledge subtraction, or GenKnowSub. (b) represents the dynamic task adaptation stage in a model layer, where the Arrow algorithm selects and combines the most relevant task-specific modules for each input token.}
    \label{fig:approach_overview}
\end{figure*}

In this paper, we adopt LoRA as the PEFT module and Arrow \citep{ostapenko2024towards} as the routing function. We choose Arrow for its ability to dynamically route each input token—rather than the entire input—to the most relevant task-specific modules in a post-hoc manner, without requiring additional training.
We hypothesize that \textit{redundant general knowledge within task-specific modules hampers generalization}. To mitigate that, we build a general knowledge LoRA using a general corpus, and then subtract it from each task LoRA. We call this process \textit{GenKnowSub}, general knowledge subtraction. The Arrow algorithm then dynamically selects and integrates the most relevant LoRAs for each input token. An overview of the proposed method can be found in Figure \ref{fig:approach_overview}.

The core intuition behind GenKnowSub is that reducing redundant general knowledge while preserving essential task-specific knowledge improves the model’s effectiveness in zero-shot transfer learning. By disentangling task-specific and general-domain knowledge, we prevent redundancy and enable better adaptation. Additionally, removing redundant knowledge enhances the distinctiveness of residual modules, ensuring that routing mechanisms can more effectively select and compose appropriate modules for solving new tasks.

We evaluate our approach mainly on Phi-3 \citep{abdin2024phi3technicalreporthighly} in a large set of benchmarks across English, German, and French. Experimental results demonstrate noticeable performance gains when compared to the base and Arrow models, underscoring the effectiveness of GenKnowSub in reducing redundancy and enhancing task-specific generalization.
We further experiment with Phi-2 \citep{javaheripi2023phi} model and show how our findings extend to this model which is more English-focused with less general capabilities. 

% As LLMs continue to evolve, techniques that enhance their adaptability and efficiency will become increasingly vital, making our proposed method a valuable addition to the field.
% We introduce \textbf{General Knowledge Subtraction (GenKnowSub)}, a novel method for enhancing modularity in LLMs by removing redundant general knowledge from task-specific LoRAs. 

Here are our key contributions:
(i) We propose GenKnowSub, a novel approach for general knowledge disentanglement by subtracting a general LoRA from task-specific LoRAs. 
GenKnowSub is simple, scalable, and seamlessly adaptable, making it applicable to broader modular LLM frameworks.
(ii) We experimentally show that GenKnowSub improves the standard Arrow method and the Phi-3 baseline performance across multiple benchmarks and languages.

\section{Method}

In this work, we address zero-shot transfer learning problem, where the goal is to transfer knowledge from a multitask dataset to solve unseen tasks without requiring labeled data for further training. Modular approaches have emerged as promising solutions for this problem. These methods operate by first training task-specific modules and then combining them to solve unseen tasks. Here, we propose to use general knowledge modules to enhance modularity detailed in the following sections.

\subsection{Training Modules and General Knowledge Subtraction}
LoRA \citep{hu2021loralowrankadaptationlarge} is a PEFT method  \citep{han2024parameterefficient} that updates only a small set of low-rank trainable parameters while keeping the pre-trained model weights frozen. By training LoRA modules on a diverse set of tasks, we enable the acquisition of distinct task-specific skills. To effectively isolate the task-specific knowledge within each LoRA module, we leverage the principle of \textit{forgetting via negation} \citep{ilharco2023editing} in module-level \citep{zhang2023composing}. 
Specifically, we define \textit{Residual LoRA} as follows:
% \begingroup
% \setlength{\abovedisplayskip}{6pt}
% \setlength{\belowdisplayskip}{6pt}
\begin{equation}
    {LoRA}^{i}_{res} = {LoRA^i_{ts} - LoRA_{g}} 
    \label{eq:lora_subtracted}
\end{equation}
% \endgroup
where \(LoRA^i_{ts}\) denotes the module trained on task \(i\) and \(LoRA_{g}\) represents the general knowledge module. 
We name this approach as \textbf{GenKnowSub} representing general knowledge subtraction. 

We hypothesize that fine-tuning the model with LoRA on even a small Wikipedia-like dataset with a causal language modeling objective could act as a bridge or a flashback for the model, bringing forth the general knowledge it acquired during pre-training. This allows the LoRA module to represent broader linguistic and factual knowledge embedded in the base model. 
This knowledge is redundant since the base model already contains it. Further, we assume that task-specific modules include some of these redundant knowledge alongside their specific functionality. Consequently, GenKnowSub effectively removes unnecessary general knowledge influence, isolating the unique task-specific characteristics essential for solving new unseen tasks.
% By subtracting the general knowledge component, we refine the task-specific modules to better encapsulate their distinct characteristics. Subsequently, these refined modules are utilized in the Dynamic Task Adaptation stage to enhance the model's ability to generalize to unseen tasks.

\subsection{Dynamic Task Adaptation}
To enhance the adaptation to unseen tasks, we employ the Arrow routing algorithm introduced in \citet{ostapenko2024towards}, which dynamically selects the \(k\) best task-specific modules for each input token in each layer and integrates them to construct an optimal LoRA module for solving unseen tasks, based on the \textit{learning via addition} principle \citep{ilharco2023editing}. Arrow computes the SVD of each LoRA, extracts the top right singular vector as a prototype, and projects input tokens onto it. The top \(k\) coefficients are selected, softmax-normalized, and others set to zero.

% To avoid an excessive number of specialized modules, we utilized the clustered Flan dataset proposed by \citep{ostapenko2024towards}. This dataset was constructed using a Model-Based Clustering (MBC) approach, where independent LoRA modules were initially trained for each individual task. The representations of these LoRA modules were then clustered using the K-means algorithm. We assume that the clustering within this dataset effectively captures the relationships between tasks, regardless of the base model on which the LoRAs were trained. This assumption allows us to dedicate a single LoRA module reflecting specific knowledge for all the task within a cluster, thereby reducing the number of modules required.

Formally, we define the computed LoRA module in each layer of the model for each input token as:
\begingroup
\setlength{\abovedisplayskip}{6pt}
\setlength{\belowdisplayskip}{6pt}
\begin{equation}
    {LoRA}^{l}_{t} = 
    \sum^{n}_{i}{c^{i,l}_{t} LoRA^{i,l}_{res}} 
    \label{eq:lora_token_layer}
\end{equation}
\endgroup
where \(n\) is the number of trained task-specific modules, \(LoRA^{i,l}_{res}\) represents the residual LoRA trained on task \(i\) within layer \(l\) of the model, and \(c^{i,l}_{t}\) indicates the importance of \(LoRA^{i,l}_{res}\) for the input token \(t\), which is calculated using the Arrow algorithm based on the input in a zero-shot manner.

Given \(LoRA^l_{t}\), the forward path for token \(t\) within layer \(l\) of the model is formulated as:
$y^l_t = W^l_0 x^l_t + B^l_{t} A^l_{t} x^l_t$ 
% \end{equation}
where \(W^l_0\in \mathbb{R}^{d \times k}\) denotes the base model weights in layer \(l\), \(x^l_{t}\in \mathbb{R}^k\) is the input representation of token \(t\) entering layer \(l\), \(A^l_t\in \mathbb{R}^{r \times k}\), \(B^l_t\in \mathbb{R}^{d \times r}\) are the corresponding LoRA parameter matrices associated with \(LoRA^l_{t}\), and $r \ll \min(d, k)$ is the rank of low-rank decomposition. 
Figure~\ref{fig:approach_overview} shows the overview of our proposed framework, including \textit{Training the Modules}, \textit{General Knowledge Subtraction (GenKnowSub)}, and \textit{Dynamic Task Adaption} stages.

\begin{table*}[ht!]
\begin{center}
{
\small
\resizebox{\textwidth}{!}{%
\begin{tabular}{ll|ccccccccc|c}
\toprule
{\bf Method} & {\bf Setting}& {\bf PIQA} & {\bf BOOLQ} & {\bf SWAG} & {\bf HSWAG} & {\bf ARC-E} & {\bf ARC-C} & {\bf WG} & {\bf OQA} & {\bf BBH} & {\bf Avg} \\ 
% \midrule
\midrule
{\bf Phi-3}  &                   
&78.24  &81.47    &68.99  &73.59  &71.75                                &44.48  &\textbf{65.98}  &42.80  &42.83  &63.35  \\
\bf Shared  &                        
&80.00  &63.39  &71.00  &72.16  &78.77                                  &49.83  &54.46  &45.40  &41.21  &61.80  \\
\bf Mean Norm  &                        
&78.02  &80.89  &71.90  &72.53  &71.58                                  &44.48  &56.83  &44.00  &40.07  &62.25  \\
\bf Arrow  &                        
&\underline{80.20}  &80.00  &68.95  &71.89  &80.53                                  &53.85  &\textbf{65.98}  &47.40  &41.23  &65.56  \\
\midrule
\multirow{4}{*}{\bf GenKnowSub} & 
$En$
&80.20  &81.96  &70.00  &73.36  &\underline{82.11}  
&53.85  &\underline{64.72}  &48.40  &43.30  &66.43  \\
 & $De$
&\textbf{80.30} &82.01 &\textbf{73.30}  &72.79  &81.75  
&54.85 &63.30 &\textbf{49.80}  &42.04  &66.68  \\
 & $Fr$
&78.78 & \underline{82.11} & 71.64 & \textbf{74.02} & 81.75 
&\textbf{57.19} & 64.40 & 49.00 & \textbf{44.40} & \underline{67.03}  \\
 & $Avg$
&80.03  &\textbf{82.45}  &\underline{72.70}  &\underline{73.45}  &\textbf{82.28}  
&\underline{55.85}  &64.64  &\underline{49.60}  &\underline{43.51}  &\bf 67.17  \\
\bottomrule
\end{tabular}
}
}
\caption{\small Comparison of accuracy across different methods using Phi-3 as the base model on some \textbf{English} reasoning datasets in a zero-shot setting. The ``Setting'' column refers to the general knowledge LoRA used for subtraction in GenKnowSub— e.g., 'En' indicates subtraction of the English general LoRA.}
\label{tab:zero-shot}
\end{center}

\end{table*}

\section{Experimental Setup and Results}
Here, we first discuss some specifications of our experimental setup including how we build our LoRA modules, and then overview the results.
\subsection{Constructing Task-Specific Modules}
\label{sec3.1}
As stated earlier, the initial step for our proposed method, GenKnowSub, involves training modules, each tailored to a specific task or functionality. To avoid an excessive number of specialized modules, we utilize clustered Flan dataset \citep{pmlr-v202-longpre23a} proposed by \citep{ostapenko2024towards}, which contains only English tasks. This dataset was constructed using a model-based clustering approach, where independent LoRAs were initially trained for each task and then clustered using the K-means algorithm. We assume that the clustering within this dataset effectively captures the relationships between tasks, regardless of the base model on which the LoRAs are trained. This assumption allows us to dedicate a single LoRA for each cluster of tasks, thereby reducing the number of experts required without compromising task-specific performance. 

We select Phi-3-mini-4k-instruct \citep{abdin2024phi3technicalreporthighly}, a 3.8-billion-parameter model, for its strong instruction-following abilities and reasonable multilingual proficiency. To address hardware limitations, we use only 20\% of each cluster’s data (\(\sim\)2,000 samples) to improve training efficiency.
The details regarding the setup and hyperparameters can be found in Appendix \ref{app:implementation}.

\subsection{Creating General Knowledge Modules}
% In order to obtain a LoRA module that effectively captures the general knowledge of a language, we train LoRA on a Wikipedia corpus using a Causal Language Modeling (CLM) objective. A LoRA module trained on Wikipedia is expected to encode broad, formal, and encyclopedic knowledge, much of which is already embedded in the underlying pre-trained model. 

To obtain a module that effectively captures the general knowledge of a language, we train LoRAs on small Wikipedia corpora using causal language modeling. We select three higher-resource languages for Phi-3 model: {English}, {French}, and {German}. We assess their impact on GenKnowSub through various combinations across multilingual zero-shot benchmarks. For the Equation~\eqref{eq:lora_subtracted}, we define \(LoRA_g\) as follows:
$LoRA_{g} = \{LoRA_{en}, LoRA_{de}, LoRA_{fr}, LoRA_{avg}\}$. 
Each LoRA is trained on 5,000 Wikipedia segments per language (details in Appendix \ref{app:implementation}), with $LoRA_{avg}$ as their average.

\subsection{Results}
We compare GenKnowSub against a range of baselines to contextualize its performance. These include the base \textit{Phi-3} model, \textit{Arrow} (Ostapenko et al., 2024), and two additional ablations introduced in this work: (i) \textit{Shared}, a single LoRA trained on the full multitask subset (20\% of each cluster); and (ii) \textit{Mean Normalization}, where the average of all task-specific LoRA modules is subtracted from each individual module as an alternative means of removing redundant information. While other recent approaches—such as Poly \citep{ponti-etal-2023-combining} and MHR \citep{caccia2023multihead}—offer additional insight into modular routing, they rely on joint training of experts and routing mechanisms, and are therefore not directly comparable in our setup.

Table \ref{tab:zero-shot} presents the performance on nine \textbf{English reasoning benchmark datasets}, including PIQA \citep{Bisk2019PIQARA}, BoolQ \citep{clark-etal-2019-boolq}, SWAG \citep{zellers-etal-2018-swag}, HellaSwag \citep{zellers-etal-2019-hellaswag}, ARC-Easy and ARC-Challenge \citep{Clark2018ThinkYH}, WinoGrande \citep{10.1145/3474381}, BIG-Bench Hard \citep{suzgun-etal-2023-challenging}, and OpenBookQA \citep{mihaylov-etal-2018-suit}.
We evaluate the impact of the different configurations of GenKnowSub on dynamic task adaptation.

As shown in Table~\ref{tab:zero-shot}, modular approaches significantly outperform the non-modular baseline. Specifically, both \textit{Arrow} and \textit{GenKnowSub} improve substantially over the \textit{Shared} baseline, confirming the effectiveness of modularity in parameter-efficient transfer. Beyond this, \textit{GenKnowSub} further enhances performance over \textit{Arrow} by removing redundant general knowledge from task-specific modules. When using the average general LoRA for subtraction, GenKnowSub achieves a consistent gain of 1.6\% over Arrow. Notably, this performance is not matched by the \textit{Mean Normalization} baseline, which naively subtracts the average of task-specific modules and yields inconsistent improvements. This highlights that our targeted subtraction of language-informed general knowledge is key to the gains observed. Finally, the effectiveness of GenKnowSub is also evident when subtracting individual language-specific LoRAs, suggesting that even language-tied general modules encode broadly shared knowledge. 

To evaluate the effectiveness of GenKnowSub beyond multiple-choice settings, we conduct experiments on the Super-Natural Instructions (SNI) dataset~\citep{wang-etal-2022-super}, a large and diverse benchmark for open-ended generation tasks. We specifically select SNI to maintain a strict zero-shot setting, consistent with our earlier evaluations on multiple-choice tasks. Additionally, SNI has been widely used to assess generalization ability in modular approaches, including in the original Arrow~\citep{ostapenko2024towards} work. Evaluating on 10{,}000 randomly sampled test examples covering 119 open-ended tasks, \textit{GenKnowSub} achieves a Rouge-L score of 46.91, outperforming the base \textit{Phi-3} model (42.85), \textit{Arrow} (45.44), the \textit{Shared} baseline (34.48), and the \textit{Mean Normalization} baseline (43.07). These results further demonstrate the generality and effectiveness of our approach in open-ended generation tasks under a zero-shot setting.

\begin{table}[h]
\begin{center}{
\small
\resizebox{\linewidth}{!}{%
\begin{tabular}{cll|cccc|c}
\toprule

& {\bf Method} & {\bf Setting} & {\bf HSWAG} & {\bf ARC-C} & {\bf XNLI} & {\bf MMLU} & {\bf Avg}\\ 
% \bottomrule
\toprule

 \multirow{8}{*}{\rotatebox[origin=c]{90}{\emph{\textbf{German}}}} &
 \bf Phi-3 & 
   &\textbf{52.48} & 36.24 & 36.02 & 33.82 & 39.64 \\

   & \bf Shared &  
   &49.75 & 38.93 & 43.00 & 36.00 & 41.92 \\

 & \bf Mean Norm &  
   &51.00 & 36.91 & 33.67 & 33.50 & 38.77 \\

 & \bf Arrow &  
   &48.58 & 40.94 & 43.45 & 35.40 & 42.09 \\
 \cmidrule{2-8}

   & \multirow{4}{*}{\bf GenKnowSub} 
   & $En$ 
   &\underline{51.16} & 40.60 & 50.14 & 36.85 & 44.69 \\

   & & $De$  
   &50.58 & \textbf{42.95} & \underline{50.42} & 37.00 & \underline{45.24}  \\

   & & $Fr$  
   &50.58 & \underline{42.62} & 49.17 & \underline{37.17} & 44.88 \\

   & & $Avg$  
   &51.08 & \underline{42.62} & \textbf{52.33} & \textbf{37.92} & \bf 45.99  \\
  
\bottomrule
\toprule

 \multirow{8}{*}{\rotatebox[origin=c]{90}{\emph{\textbf{French}}}} &
 \bf Phi-3 & 
   &\underline{57.67} & 34.56 & 50.75 & 33.33 & 44.08 \\

    & \bf Shared &  
   &57.08 & 40.27 & 50.17 & 35.33 & 45.71 \\

   & \bf Mean Norm &  
   &57.42 & 35.91 & 52.42 & 33.25 & 44.75 \\
   
   & \bf Arrow &  
   &55.33 & 41.61 & 44.38 & 34.79 & 44.02 \\
 \cmidrule{2-8}

   & \multirow{4}{*}{\bf GenKnowSub} 
   & $En$ 
   &56.08 & 41.95 & 50.66 & \textbf{36.69} & 46.34 \\

   & & $Fr$  
   &\textbf{57.83} & \textbf{42.95} & \textbf{53.65} & 36.13 & \bf 47.64  \\

   & & $De$  
   &56.42 & 42.28 & 46.33 & \underline{35.58} & 45.15 \\

   & & $Avg$  
   &57.08 & \underline{42.62} & \underline{52.92} & \underline{35.58} & \underline{47.05}  \\

\bottomrule
\end{tabular}
}
}
\end{center}
\caption{\small Performance comparison of different methods with Phi-3 as the base model in a zero-shot setting for \textbf{German} and \textbf{French} languages. Various configurations of GenKnowSub are evaluated, with accuracy as the reported metric.}
\label{tab:zero-shot-ml}
\end{table}

To evaluate GenKnowSub across \textbf{non-English languages}, we use XNLI \citep{conneau-etal-2018-xnli}, the translated versions of the HellaSwag, MMLU \citep{hendrycks2021measuring}, and ARC-Challenge datasets provided by \citep{Lai2023OkapiIL}. Table~\ref{tab:zero-shot-ml} shows that \textit{GenKnowSub} consistently achieves the best performance across both German and French benchmarks. In contrast, \textit{Arrow} exhibits inconsistent results—outperforming \textit{Shared} and \textit{Mean Norm} in German, but falling behind them in French—highlighting its variability across languages. GenKnowSub surpasses all baselines in both settings, with its strongest configuration (average subtraction) improving over Arrow by 3.9\% in German and 3.6\% in French. These results confirm the generality and robustness of our approach, and reinforce the findings from the previous experiments on the English benchmark datasets: removing shared general knowledge from task-specific modules before task adaptation leads to more effective zero-shot transfer across languages.

A key factor in cross-lingual transfer learning is the base model’s ability to encode at least a minimal level of multilinguality. To assess its impact more precisely, we run an additional experiment using \textbf{Phi-2}, which is weaker than Phi-3 in both multilingual and instruction-following capability. Following \citep{ostapenko2024towards}, we use unquantized Phi-2 here, with task modules trained on the full task cluster data. As shown in Table \ref{tab:phi2-en} in Appendix, GenKnowSub, after subtracting the English general knowledge module, improves performance on English benchmark datasets in a zero-shot setting, increasing the average score by 1.1\%. However, in German and French experiments (Table \ref{tab:phi2-ml} in Appendix), both the base Phi-2 model and its combination with Arrow perform poorly—around 13\% lower than Phi-3 in a similar setting—due to Phi-2’s weak multilingual capabilities. Consequently, GenKnowSub achieves only comparable results, underperforming Arrow by 0.3\%. These findings further confirm that our approach can enhance performance, provided the base model has at least a minimal level of multilinguality. Additional details and results are provided in Appendix \ref{app:phi2}.

% One key observation is the improvement the Arrow method achieved on the French and German benchmark datasets. This highlights the critical role of learning task-specific knowledge and optimally combining these modules in zero-shot transfer tasks, regardless of the language in which the task-specific modules were trained. Although our modules were trained on an English multi-task dataset, the effective combination of these modules using Arrow yielded strong performance on unseen tasks in French and German, demonstrating the language-agnostic nature of solving the tasks.

\section{Conclusion}
In this work, we propose a modular approach to zero-shot transfer learning, leveraging task-specific and general knowledge modules to enhance adaptability to unseen tasks. Our method first isolates task-relevant representations through GenKnowSub, then dynamically adapts these modules using the Arrow routing algorithm. By minimizing redundancy in task representations, our approach improves both efficiency and transferability.
We demonstrate that applying GenKnowSub prior to task adaptation improves generalization in zero-shot settings for both Phi-3 and Phi-2 models across both multiple-choice and open-ended generation tasks.
Our results show that this method not only enhances performance in monolingual tasks but also facilitates effective cross-lingual transfer when the language is highly present in the base model. 
% Furthermore, we find that general knowledge is largely shared across languages, allowing representations from high-resource languages to enhance transferability to low-resource languages by reducing redundancy in each module.
Future work includes exploring alternative task adaptation methods, extending our approach to additional languages, especially low-resource ones, and testing it on other models.

\section*{Limitations}
One limitation of this study is the restricted scope of model evaluation. Due to hardware constraints—such as limited GPU VRAM and slower processing speeds—we limited our experiments to two models: Phi-3 and Phi-2. These constraints precluded testing on larger or more diverse models, thereby limiting the breadth of our analysis. Additionally, although we included multilingual evaluations, our focus remained on high-resource languages (e.g., English, French, German), and we did not investigate performance in low-resource settings. Future work should aim to expand the evaluation to a broader range of models, tasks, and languages to further assess the generality of the proposed approach.

% These limitations suggest several directions for future research. Evaluating our approach on a wider set of tasks, testing it across various LLM architectures, and expanding its applicability to lower-resource languages would provide deeper insights into its effectiveness. By addressing these challenges, we can further refine our method and enhance its usability in diverse NLP applications.

\section*{Ethics}
Our research utilizes publicly available datasets and pre-trained models, ensuring compliance with ethical data usage practices and avoiding the use of private, proprietary, or personally identifiable information. All models and associated code will be made publicly available under permissive licenses, promoting accessibility, reproducibility, and unrestricted use for research and application development. However, we acknowledge that pre-trained language models (PLMs) and large language models (LLMs) have been shown to exhibit biases, as highlighted in prior work \citep{Liang2021TowardsUA, 10337300}. Users should be mindful of these limitations when applying such models in practice. Our work does not introduce additional fairness or privacy concerns.

% Bibliography entries for the entire Anthology, followed by custom entries
% \bibliography{anthology,custom}
% Custom bibliography entries only
\bibliography{acl_latex}

\appendix
\begin{table*}[!t]
\begin{center}
{
\small
\resizebox{\textwidth}{!}{%
\begin{tabular}{ll|ccccccccc|c}
\toprule
{\bf Method} & {\bf Setting}& {\bf PIQA} & {\bf BOOLQ} & {\bf SWAG} & {\bf HSWAG} & {\bf ARC-E} & {\bf ARC-C} & {\bf WG} & {\bf OQA} & {\bf BBH} & {\bf Avg} \\ 
\midrule
\midrule
{\bf Phi-2}  &                   
&78.99 &81.16 &63.50 &66.75 &82.11 &53.51                                 &56.51 &44.00 &48.00 &63.84    \\
\bf Arrow  &                        
&79.65 &81.13 &65.75 &66.41 &83.38 &54.84                                 &60.85 &48.60 &54.75 &65.15    \\  
\midrule
\multirow{2}{*}{\bf GenKnowSub} & $En$
&79.97 &80.12 &65.58 &66.75 &84.38 &54.51                                 &61.24 &49.80 &54.00 &\bf 66.26    \\
%  & $- De$
% &80.30 &82.01 &73.30  &72.79  &81.75  
% &54.85 &63.30 &49.80  &42.04  &66.68  \\
%  & $- Fr$
% &78.78 & 82.11 & 71.64 & 74.02 & 81.75 
% &57.19 & 64.40 & 49.00 & 44.40 & \underline{67.03}  \\
%  & $- En - De$
% &80.14 &81.13  &72.80  &72.77  &82.11  
% &55.85 &64.64  &47.40  &42.93  &66.64  \\
%  & $- En- Fr$
% &79.49  &81.68  &73.40  &72.28  &82.63  
% &55.85  &63.61  &47.00  &44.55  &66.72  \\
%  & $- En - De - Fr$
% &79.54  &80.09  &73.40  &72.50  &81.93  
% &55.52  &63.46  &46.20  &44.35  &66.33  \\
 & $Avg$
&80.47 & 78.47 & 66.10 & 67.96 & 84.03 
&56.19 & 60.69 & 47.80 & 54.00 & \underline{66.19} \\
\bottomrule
\end{tabular}
}
}
\caption{\small Comparison of accuracy across different methods using Phi-2 as the base model on English datasets in a zero-shot setting, with Accuracy as the reported metric.}
\label{tab:phi2-en}
\end{center}

\end{table*}
\begin{table}[!h]
\begin{center}{
\small
\resizebox{\linewidth}{!}{%
\begin{tabular}{cll|cccc|c}
\toprule

& {\bf Method} & {\bf Setting} & {\bf HSWAG} & {\bf ARC-C} & {\bf XNLI} & {\bf MMLU} & {\bf Avg}\\ 
\bottomrule
\toprule

 \multirow{4}{*}{\rotatebox[origin=c]{90}{\emph{\textbf{German}}}} &
 \bf Phi-2 & 
   &28.78 & 23.84 & 34.50 & 24.19 & 27.83 \\

   & \bf Arrow &  
   &28.75 & 24.83 & 32.50 & 26.91 & \bf 28.25 \\
 \cmidrule{2-8}

   & \multirow{2}{*}{\bf GenKnowSub} & $En$ 
   &28.33 & 24.55 & 33.33 & 24.91 & 27.78 \\

   & & $Avg$  
   &28.42 & 23.49 & 34.33 & 25.50 & \underline{27.93}   \\
  
\bottomrule
\toprule

 \multirow{4}{*}{\rotatebox[origin=c]{90}{\emph{\textbf{French}}}} &
 \bf Phi-2 & 
   &33.33 & 26.84 & 34.16 & 24.77 & 29.77 \\

   & \bf Arrow &  
   &32.91 & 27.51 & 34.16 & 25.68 & \bf 30.06 \\
 \cmidrule{2-8}

   & \multirow{2}{*}{\bf GenKnowSub} & $En$ 
   &33.50 & 25.50 & 31.83 & 26.11 & 29.23 \\

   & & $Avg$  
   &32.25 & 24.16 & 37.50 & 25.00 & \underline{29.73}   \\

\bottomrule
\end{tabular}
}
}
\end{center}
\caption{\small Performance comparison of different methods using Phi-2 as the base model on multilingual datasets in a zero-shot setting for German and French, with Accuracy as the reported metric.}
\label{tab:phi2-ml}
\end{table}

\section{Implementation Details}
\label{app:implementation}
\subsection{Base Model}
We utilize Phi-3-mini-4k-instruct \citep{abdin2024phi3technicalreporthighly} with 4-bit quantization to reduce memory usage while maintaining strong performance. We selected this model due to its exceptional instruction-following capabilities and its acceptable multilingual proficiency. Additionally, with only 3.8 billion parameters, Phi-3-mini strikes an effective balance between model size and performance, allowing for efficient fine-tuning and deployment in resource-constrained environments while still demonstrating competitive reasoning and generalization abilities. 

\subsection{PEFT Structure}
As the PEFT structure, we employ LoftQ \citep{li2024loftq} with a rank of \(r=4\). LoftQ extends LoRA by integrating low-rank adaptation directly into the quantization process, thereby optimizing both fine-tuning and inference through rank-wise quantization, which minimizes precision loss while updating quantized model weights. We applied our PEFT modules to both the QKV components (concatenation of Query, Key, and Value matrices in the self-attention block) and the output projection layer of the multi-head attention.

\subsection{Arrow Routing}
\label{app:arrow}
To incorporate the Arrow routing algorithm, we implemented it from scratch using PyTorch \citep{NEURIPS2019_bdbca288} and the PEFT library from HuggingFace \citep{wolf-etal-2020-transformers}. We trained 10 task-specific modules, and selected the 3 best modules for each input token in each layer of the model to be combined.

\subsection{Module Training}
\label{app:training}
To train the LoRA modules representing general knowledge, we fine-tuned the model on a Wikipedia dataset using a causal language modeling objective on a single Quadro RTX 6000 GPU. We ensured consistency across languages by sampling exactly 5,000 segments per language from the Hugging Face wikimedia/wikisource \citep{wikidump} dataset. Each segment contained 512 words, which were split into a 507-word prompt and a 5-word completion. LoRA modules were fine-tuned using the same supervised setup for all languages. This uniform approach ensured that the amount, structure, and formatting of training data were identical for each language, mitigating any length- or volume-based bias.

For task-specific LoRA modules, we employed a supervised fine-tuning approach. Given the presence of relatively long examples in our dataset, we set the maximum sequence length to 4000 tokens to accommodate the full input structure.

Both the task-specific and general knowledge LoRA modules were trained for one epoch with a learning rate of \(1e{-4}\), using cosine scheduling with a warmup start. To stabilize training, we applied gradient clipping. Additionally, to optimize memory efficiency, we utilized the Paged AdamW 8-bit optimizer \citep{dettmers2022bit}, a quantized variant of AdamW \citep{loshchilov2018decoupled}, designed to reduce GPU memory consumption. 

The batch size was set to 16 for training general knowledge modules and 1 for task-specific modules, due to the long input lengths. To further improve memory efficiency, we applied gradient checkpointing and gradient accumulation, enabling support for larger batch sizes when training task-specific modules.

\section{Experiments on Phi-2}
\label{app:phi2}
\subsection{Implementation Details}
For conducting experiments using Phi-2, the baseline in \citet{ostapenko2024towards}, we used the LoRA modules they trained to evaluate their proposed routing algorithm, Arrow (the implementation details of Arrow are provided in Appendix \ref{app:arrow}). Specifically, we used task LoRA modules trained by \citet{ostapenko2024towards}, available on Hugging Face.\footnote{Library of LoRAs: \url{https://huggingface.co/zhan1993/mbc_library_phi2_icml}} These modules are obtained by fine-tuning Phi-2 on clustered Flan datasets (explained in Section \ref{sec3.1}) and are provided in PyTorch Lightning format.\footnote{PyTorch Lightning: \url{https://github.com/Lightning-AI/pytorch-lightning}} Since our models are trained using the PEFT library,\footnote{PEFT: \url{https://github.com/huggingface/peft}} we converted the existing LoRA weights into the PEFT format to ensure compatibility. Our implementation loads expert weights following the PEFT framework, and, consistent with their setup, the Phi-2 experiment weights remain unquantized.

Additionally, we obtained general knowledge LoRA modules by fine-tuning Phi-2 in a setup aligned with the task-specific LoRA modules. The training process was similar to that of the Phi-3 version, as detailed in Appendix \ref{app:training}. 
% However, since Phi-2 underperforms compared to Phi-3 in multilingual and instruction-following tasks, we chose Phi-3 to better harness the base model's transferability. 
\subsection{Resutls}
We demonstrated that with a sufficiently strong multilingual base model, we can effectively leverage its multilingual capabilities to generalize better to unseen tasks across different languages. The Phi-2 experiments further highlight the importance of base model strength and knowledge. As shown in Table \ref{tab:phi2-en}, GenKnowSub, with subtracting the English general knowledge module, outperforms other settings, whereas averaging modules across different languages is less effective than using English alone. Additionally, Table \ref{tab:phi2-ml} shows that all settings, including Phi-2 and Arrow, perform poorly in German and French. The improvement of our method on the English zero-shot dataset, along with its performance in the multilingual setting, demonstrates that our method can significantly enhance results—provided the base model exhibits at least a minimal level of cross-lingual capability.

\end{document}